\documentclass[10pt,twocolumn,letterpaper]{article}

\usepackage[pagenumbers]{cvpr} %

\usepackage[noend]{algpseudocode}
\usepackage{algorithm}
\usepackage{amsmath}
\usepackage[dvipsnames]{xcolor}

\newcommand{\todocite}[1][]{%
    \ifx\relax#1\relax
        {\color{red}(cite )}%
    \else
        {\color{red}(cite #1 )}%
    \fi
}

\usepackage{amsfonts}
\usepackage{amsmath}
\usepackage{amssymb}
\usepackage{booktabs}
\usepackage{multirow}
\usepackage[font=footnotesize]{caption}
\usepackage{cuted}
\usepackage{diagbox}
\usepackage{float}
\usepackage{graphicx}
\usepackage{microtype}
\usepackage{nicefrac}
\usepackage{pifont}
\usepackage{times}
\usepackage{xcolor}
\usepackage[accsupp]{axessibility}  %

\definecolor{cvprblue}{rgb}{0.21,0.49,0.74}
\usepackage[pagebackref,breaklinks,colorlinks,citecolor=cvprblue]{hyperref}
\usepackage[capitalize]{cleveref}

\def\paperID{16570} %
\def\confName{CVPR}
\def\confYear{2024}

\newcommand{\methodname}{Fixed Point Diffusion Model}
\newcommand{\shortname}{FPDM}

\newcommand{\ppx}[2]{\frac{\partial #1}{\partial #2}}

\newcommand{\fL}{\mathcal{L}}

\title{Fixed Point Diffusion Models}

\author{Xingjian Bai
\thanks{Equal Contribution.} 
\\
University of Oxford \\
{\tt\small xingjianbai@gmail.com}
\and
Luke Melas-Kyriazi
\footnotemark[1]
\thanks{Corresponding author.}
\\
University of Oxford \\
{\tt\small lukemk@robots.ox.ac.uk} \\
}

\begin{document}
\maketitle
\begin{strip}
\centering
\includegraphics[width=1.0\linewidth]{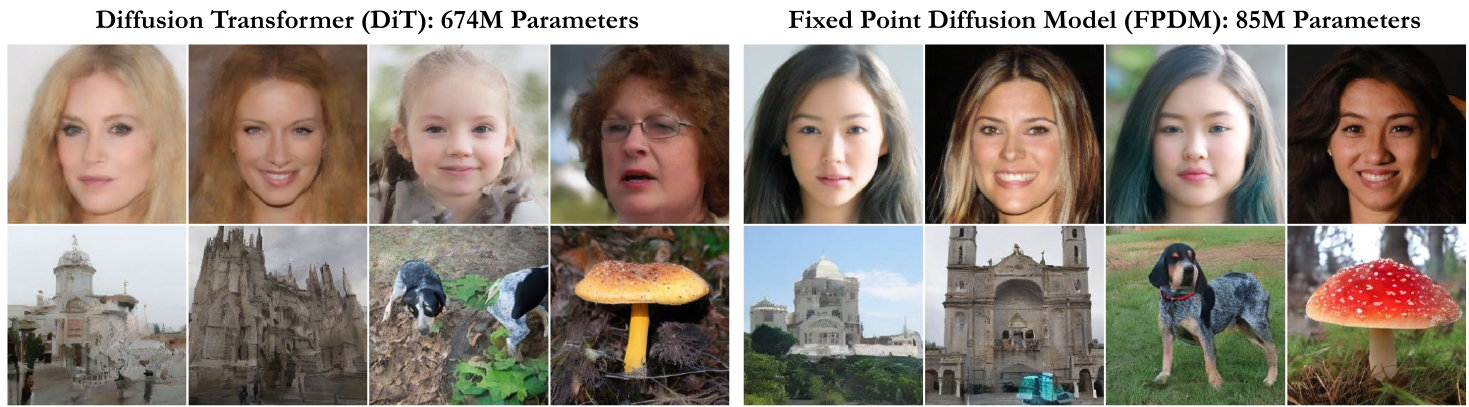}
\captionof{figure}{
\textbf{\methodname{} (\shortname)} is a novel and highly efficient approach to image generation with diffusion models. \shortname{} integrates an implicit fixed point layer into a denoising diffusion model, converting the sampling process into a sequence of fixed point equations. Our model significantly decreases model size and memory usage while improving performance in settings with limited sampling time or computation. 
We compare our model, trained at a 256 $\times$ 256 resolution against the state-of-the-art DiT~\cite{william22scalable} on four datasets (FFHQ, CelebA-HQ, LSUN-Church, ImageNet) using compute equivalent to $20$ DiT sampling steps. 
\shortname{} (right) demonstrates enhanced image quality with 87\% fewer parameters and 60\% less memory during training. 
}%
\vspace{3mm}
\label{fig:splash}
\end{strip}
\begin{abstract}
We introduce the Fixed Point Diffusion Model (\shortname), a novel approach to image generation that integrates the concept of fixed point solving into the framework of diffusion-based generative modeling. 
Our approach embeds an implicit fixed point solving layer into the denoising network of a diffusion model, transforming the diffusion process into a sequence of closely-related fixed point problems. 
Combined with a new stochastic training method, this approach significantly reduces model size, reduces memory usage, and accelerates training.
Moreover, it enables the development of two new techniques to improve sampling efficiency: reallocating computation across timesteps and reusing fixed point solutions between timesteps.  
We conduct extensive experiments with state-of-the-art models on ImageNet, FFHQ, CelebA-HQ, and LSUN-Church, demonstrating substantial improvements in performance and efficiency.
Compared to the state-of-the-art DiT model~\cite{william22scalable}, \shortname{} contains 87\% fewer parameters, consumes 60\% less memory during training, and improves image generation quality in situations where sampling computation or time is limited. 
Our code and pretrained models are available at \href{https://lukemelas.github.io/fixed-point-diffusion-models/}{https://lukemelas.github.io/fixed-point-diffusion-models/}.

\end{abstract}
\section{Introduction}
\label{sec:intro}

The field of image generation has experienced significant recent advancements driven by the development of large-scale diffusion models~\cite{ho2020denoising,song2020denoising,william22scalable,pmlr-v162-nichol22a,rombach2022high,song2019generative}. Key to these advancements have been increased model size, computational power, and the collection of extensive  datasets~\cite{yu2015lsun,karras2017progressive,bai2022ffhquv,deng2009imagenet,laion400m,laion5b,gadre23datacomp}, collectively contributing to a marked improvement in generation performance.

Despite these strides, the core principles of diffusion networks have remained largely unchanged since their development~\cite{ho2020denoising}. They typically consist of a fixed series of neural network layers, either with a UNet architecture ~\cite{klaus2017unet} or, more recently, a vision transformer architecture~\cite{attentionisallyouneed,dosovitskiy2020image}. However, as diffusion models are increasingly deployed in production, especially on mobile and edge devices, their large size and computational costs pose significant challenges.

This paper introduces the \methodname{} (\shortname), which integrates an implicit fixed point solving layer into the denoising network of a diffusion model. 
In contrast to traditional networks with a fixed number of layers, \shortname{} is able to utilize a variable amount of computation at each timestep, with the amount of computation directly influencing the accuracy of the resulting solutions. 
This fixed point network is then applied sequentially, as in standard diffusion models, to progressively denoise a data sample from pure Gaussian noise. 

\shortname{} offers efficiency gains at two levels of granularity: that of individual timesteps and that of the entire diffusion process. First, at the timestep level, it provides:
\begin{enumerate}
\item A substantial reduction in parameter count compared to previous networks (87\% compared to DiT~\cite{william22scalable}).
\item Reduced memory usage during both training and sampling (60\% compared to DiT~\cite{william22scalable}).
\end{enumerate}
Second, at the diffusion process level, it provides: 
\begin{enumerate}
  \item The ability to smoothly distribute or reallocate computation among timesteps. This contrasts with all previous diffusion models, which must perform a full forward pass at every sampling timestep.
  \item The capacity to reuse solutions from one fixed-point layer as an initialization for the layer in the subsequent timestep, further improving efficiency.
\end{enumerate}
Our fixed-point network thereby delivers immediate benefits, in the form of reduced size and memory (\cref{sec:methods_network}), and further benefits when integrated into the diffusion process, in the form of increased flexibility during sampling (\cref{sec:methods_fpdm}).

To realize these benefits, it is imperative to train our models using an efficient and effective differentiable fixed-point solver. Although several implicit training methods exist in the literature~\cite{bai19deep,geng21on,fung22jacobian}, we find these existing approaches to be unstable or underperformant in our setting. Hence, we develop a new training procedure named Stochastic Jacobian-Free Backpropagation (S-JFB) (\cref{ssec:backprop}), inspired by Jacobian-Free Backpropagation (JFB)~\cite{fung22jacobian}. This procedure is stable, highly memory-efficient, and surpasses standard JFB in performance.

We demonstrate the efficacy of our method through extensive experiments (\cref{sec:experiments}) on four popular image generation datasets: LSUN-Church~\cite{yu2015lsun}, CelebA-HQ~\cite{karras2017progressive}, FFHQ~\cite{bai2022ffhquv}, and ImageNet~\cite{deng2009imagenet}. 
\shortname{} excels over standard diffusion models when computational resources during sampling are limited.
Finally, detailed analysis and ablation studies (\cref{sec:ablations}) demonstrate the efficacy of our proposed network, sampling techniques, and training methods.

\begin{figure}
\centering
\includegraphics[width=0.5\textwidth]{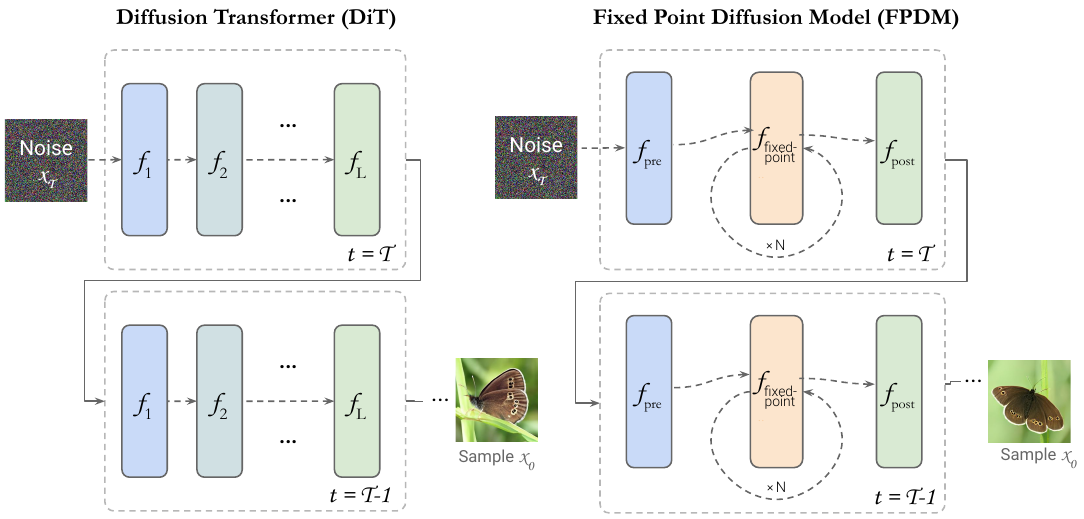}
\caption{\textbf{The architecture of FPDM compared with DiT.} FPDM keeps the first and last transformer block as pre and post processing layers and replaces the explicit layers in-between with an implicit fixed point layer. Sampling from the full reverse diffusion process involves solving many of these fixed point layers in sequence, which enables the development of new techniques such as timestep smoothing (\cref{sec:methods_smoothing}) and solution reuse (\cref{sec:methods_reuse}).}%
\vspace{-3mm}
\label{fig:arch_figure}
\end{figure}
\section{Related Work}
\label{sec:formatting}
\paragraph{Diffusion Models (DMs).}
Diffusion models~\cite{ho2020denoising,anand2022protein}, or score-based generative models~\cite{song2019generative,song2020improved}, are the source of tremendous recent progress in image generation. 
They learn to reverse a Markovian noising process using a denoiser parametrized by a neural network, traditionally a U-Net~\cite{klaus2017unet}.
The denoising paradigm can be seen as the discretization of a stochastic differential equation in a continuous domain~\cite{song2021scorebased}. 
Later work equipped DMs with different sampling methods~\cite{liu2021pseudo,watson2021learning,song2020denoising} and  applied conditional control from multiple modalities~\cite{dhariwal2021diffusion,song2020improved,nichol2021glide}. 
Recently, DMs with transformer-based architectures (DiTs) were shown to be highly effective~\cite{william22scalable}; \shortname{} builds upon the DiT architecture.

The heavy memory and computation requirements of DMs scale up quadratically with the image resolution and linearly with the number of sampling timesteps. 
To reduce training cost, LDM~\cite{rombach2022high} proposes to downsample images with a pre-trained Variational Autoencoder~\cite{autoencoder} and perform denoising in latent space. 
However, the inference cost of DMs is still considered their primary drawback.
\paragraph{Implicit Networks and Deep Equilibrium Models.}
Whereas traditional neural networks calculate outputs by performing a pass through a stack of layers, implicit neural networks define their outputs by the solutions of dynamic systems.
Specifically, Deep Equilibrium Models (DEQs)~\cite{bai19deep,torchdeq} define their output by the fixed point of an equilibrium layer $f_\theta$. 
The equilibrium state of DEQs, $z^\ast$, is equivalent to the output of an infinite-depth, weight-sharing explicit neural network:
$\lim_{k \to \infty} f_\theta(z^k) = f_\theta(z^\ast) = z^\ast.$
In its forward pass, the equilibrium state $z^\ast$ can be computed by applying solvers like Broyden's method~\cite{broyden65} or Anderson's acceleration~\cite{anderson65}. In the backward pass, one can implicitly differentiate through the equilibrium state $z^\ast$, or use one of the recently-proposed accelerated training methods~\cite{geng21on,fung22jacobian}. Applications of DEQs include optical flow~\cite{bai2022deep}, image segmentation~\cite{bai2020multiscale} and inverse imaging problems~\cite{davis2021deqinverse}. 

\paragraph{Recent Work Combining Diffusion and DEQs.}
In the past year, two works have merged DMs and DEQs. 
Differently from our proposal, these approaches tried to convert the \textit{entire diffusion process} into a \textit{single} fixed point equation. 
\cite{pokle2022deep} considers the entire diffusion trajectory as a single sample and solves for the fixed point of the trajectory, converting the sequential diffusion process into a parallel one. 
\cite{geng2023onestep} distills a pretrained diffusion model into a single-step DEQ. 
These works are exciting but come with their own drawbacks: the former is an inference-time technique that consumes significantly more memory than standard ancestral sampling, while the latter requires a pretrained diffusion model and has not scaled to datasets larger than CIFAR-10.

\section{Methods}

\subsection{Preliminaries}

\paragraph*{Implicit Neural Networks.}
The neural network layer is the foundational building block of deep learning. 
While early neural networks used only a few layers~\cite{alexnet12,lenet98}, modern networks such as large-scale transformers~\cite{attentionisallyouneed,dosovitskiy2020image} often consist of dozens of layers connected by residual blocks. 
Typically, these layers share a similar internal structure and dimensionality, with each having distinct set of parameters.
In essence, most networks are defined \textit{explicitly}: their operations are precisely defined by their layer weights. Running a forward pass always entails processing inputs with the entire set of layers.

On the other hand, \textit{implicit} models define the function or procedure to be computed by the network, rather than the exact sequence of operations. This category includes models that integrate differential equation solvers (Neural ODE/CDE/SDEs;~\cite{neuralode,neuralcde,neuralsde}), as well as models incorporating fixed point solvers (fixed point networks or DEQs;~\cite{bai19deep}). Our proposed \shortname{} belongs to this latter group.

\paragraph*{Differentiable Fixed Point Solving.}
Given a function $f$ on $X$, a fixed point solver aims to compute $x^{*} \in X$ such that $f(x^{*}) = x^{*}$. 
The computation of fixed points has been the subject of centuries of mathematical study~\cite{wallis1685treatise}, with the existence and uniqueness of a system's fixed points often proved with the Banach fixed-point theorem and its variants~\cite{istratescu1981fixed,granas2003fixed,agarwal2001fixed}. 

In our case, $f = f_{\theta}$ is a differentiable function parametrized by $\theta$, and we are interested in both solving for the fixed point and backpropagating through it.
The simplest solving method is \textit{fixed point iteration}, 
which iteratively applies $f_\theta$ until convergence to $x^{*}$. 
Under suitable assumptions, iteration converges linearly to the unique fixed point of an equilibrium system (Thrm 2.1 in ~\cite{fung22jacobian}). 
Alternative methods found throughout the literature include Newton's method, quasi-Newton methods such as Broyden's method, and Anderson's acceleration. 
In these cases, one can analytically backpropagate through $x^{*}$ via implicit differentiation~\cite{fung22jacobian}. 
However, these methods can come with significant memory and computational costs. 
Recently, a new iterative solving method denoted Jacobian-Free Backpropagation (JFB) was introduced to circumvent the need for complex and costly implicit differentiation; we discuss and extend upon this method in \cref{ssec:backprop}. 

\subsection{Fixed Point Denoising Networks} \label{sec:methods_network}

Our proposed fixed-point denoising network (\cref{fig:arch_figure}) integrates an implicit fixed-point layer into a 
diffusion transformer.
The network consists of 
three stages: 1) explicit timestep-independent preprocessing layers $f_{\text{pre}}^{(1:l)}:X \to X$,  2) a implicit timestep-conditioned fixed-point layer $f_{\text{fp}}: X \times X \times T \to X$, and 3) explicit timestep-independent postprocessing layers $f_{\text{post}}^{(1:l)}: X \to X$. 
The function $f_{\text{fp}}$ takes as input both the current fixed-point solution $x$ and a value $\tilde{x}$ called the \textit{input injection}, which is the projected output of the preceding explicit layers. 
One can think of $f_{\text{fp}}$ as a map $f_{\text{fp}}^{(\tilde{x}, t)}: X \to X$ conditional on the input injection and timestep, for which we aim to find a fixed point.
The network processes an input $x_{\text{input}}^{(t)}$ as follows:
\begin{align}
x_{\text{pre}}^{(t)} &= f_{\text{pre}}^{(1:l)}(x_{\text{input}}^{(t)}) \\
\tilde{x}^{(t)} &= \text{projection}(x_{\text{pre}}^{(t)}) \quad\qquad\textit{ input injection} \\
x^{* (t)} &= f_{\text{fp}}(x^{* (t)}, \tilde{x}^{(t)}, t) \quad \textit{via fixed point solving} \label{network} \\
x_{\text{post}}^{(t)} &= f_{\text{post}}^{(1:l)}(x^{* (t)})
\end{align}
The output $x_{\text{post}}^{(t)}$ is used to compute the loss (during training) or the input $x_{\text{input}}^{(t-1)}$ to the next timestep (during sampling). 

Whereas explicit networks consume a fixed amount of computation, this implicit network can adapt based on the desired level of accuracy or even on the difficulty of the input. In this way, it unlocks a new tradeoff between computation and accuracy. 
Moreover, since the implicit layer replaces a large number of explicit layers, it significantly decreases its size and memory consumption. 

Finally, note that our denoising network operates in latent space rather than pixel space. That is, we apply a Variational Autoencoder~\cite{autoencoder,rombach2022high} to encode the input image into latent space and perform all processing in latent space.

\subsection{Fixed Point Diffusion Models (FPDM)} \label{sec:methods_fpdm}

FPDM incorporates the fixed point denoising network proposed above into a denoising diffusion process. 

We assume the reader is already familiar with the basics of diffusion models and provide only a brief summary; if not, we provide an overview in the Supplementary Material. 
Diffusion models learn to reverse a Markovian noising process in which a sample $X_0 \sim q(X_0)$ from a target data distribution $q(X_0)$ is noised over a series of timesteps $t \in [0,T]$.
The size of each step of this process is governed by a variance schedule $\{\beta_t\}_{t=0}^{T}$ as
$
q(X_t|X_{t-1}) = \mathcal{N}(X_t; \sqrt{1-\beta_t}X_{t-1}, \beta_t \mathbf{I}).
$
where each $q(X_t|X_{t-1})$ is a Gaussian distribution.
We learn the distribution $q(X_{t-1}|X_t)$ using a network $s_{\theta}(X_{t-1}|X_t) \approx q(X_{t-1}|X_t)$, which in our case is a fixed point denoising network. The generative process then begins with a sample from the noise distribution $q(X_T)$ and denoises it over a series of steps to obtain a sample from the target distribution $q(X_0)$. 

The primary drawback of diffusion models as a class of generative models is that they are relatively slow to sample. As a result, during sampling, it is very common to only use a small subset of all diffusion timesteps and take correspondingly larger sampling steps; for example, one might train with 1000 timesteps and then sample images with as few as $N=5, 10$, or $20$ timesteps. 

Naturally, one could replace the explicit denoising network inside a standard diffusion model with a fixed point denoising network, and make no other changes; this would immediately reduce model size and memory usage, as discussed  previously. 
However, we can \textit{further} improve efficiency during sampling by exploiting the fact that we are solving a \textit{sequence} of related fixed point problems across all timesteps, instead of a single one. We present two opportunities for improvement: smoothing/reallocating computation across timesteps and reusing solutions. 

\paragraph*{Smoothing 
 Computation Across Timesteps.}\label{sec:methods_smoothing}
\begin{figure}[t]
  \centering
   \includegraphics[width=1.0\linewidth]{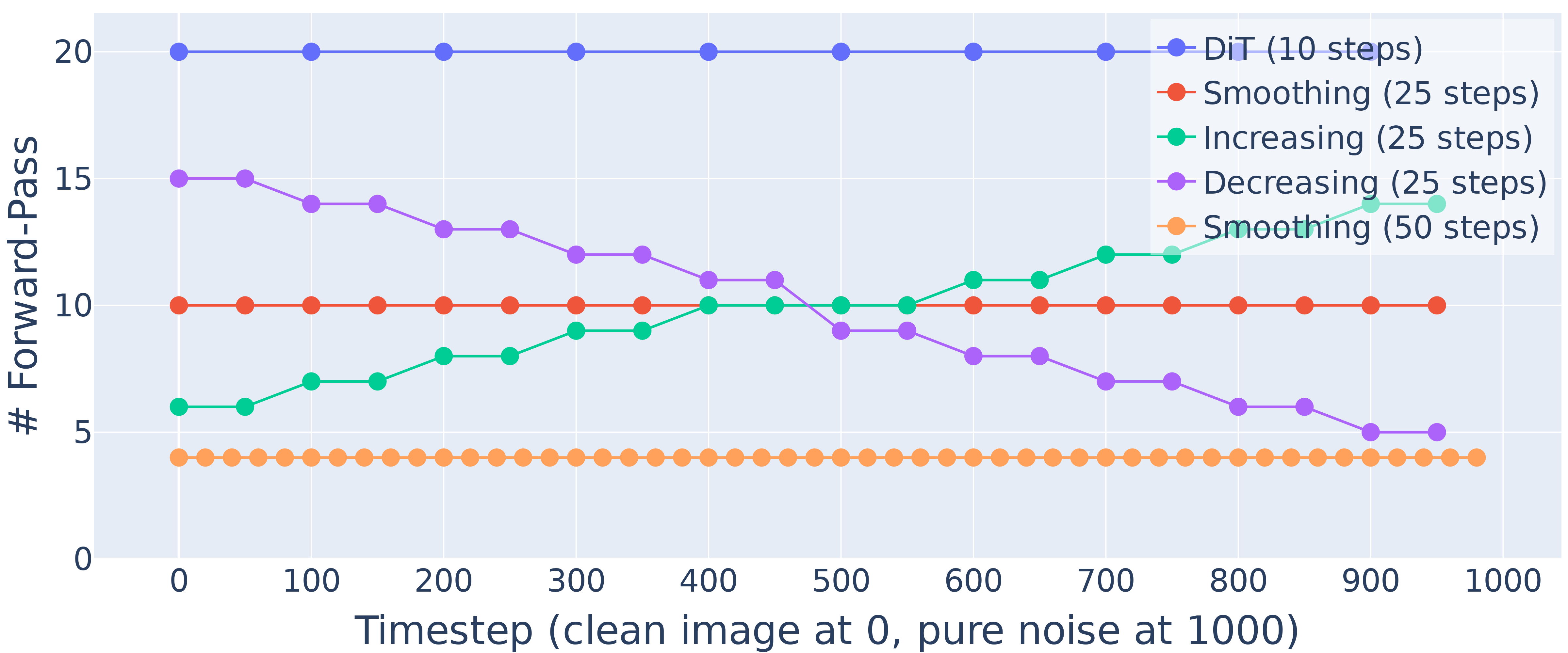}
   \caption{
   \textbf{Illustration of Transformer Block Forward Pass Allocation in \shortname{} and DiT.}
   Since DiT has to perform full forward passes at each timestep, under limited compute, it can only denoise at a few timesteps with large gaps. 
   \shortname{} achieves a more balanced distribution through smoothing, thereby reducing the discretization error. 
   Additionally, \shortname{} offers the flexibility to adjust forward pass allocation per timestep with heuristics like \textit{Increasing} and \textit{Decreasing}. Refer to Section~\ref{sec:methods_smoothing} for details.}
   \vspace{-3mm}
   \label{fig:table1}
\end{figure}

The flexibility afforded by fixed-point solving enables us to allocate computation between timesteps in a way that is not possible with traditional diffusion models. For a given computational budget for sampling, we can 
reduce the the number of forward passes (i.e. number of fixed point iterations) used per timestep in order to use more timesteps over the sampling process (see \cref{fig:table1}). In other words, our implicit model can effectively ``smooth out'' the computation over more timesteps. By contrast, with explicit models such as DiT, the amount of computation directly determines the number of timesteps, since a full forward pass is needed at each timestep. Indeed, we find that when the amount of compute is relatively limited, it is highly beneficial to smooth out the compute among more timesteps than would be done with a traditional model. The effectiveness of smoothing is shown empirically in section~\cref{sec:discussion_smoothing}.

\paragraph*{Reallocating Computation Across Timesteps}
Beyond smoothing out computation over timesteps, \shortname{} enables one to vary the number of forward passes at each timestep, thereby dynamically controlling the solving accuracy at different stages of the denoising process. 
This capability enables the implementation of diverse heuristics. 
For example, we can allocate a greater number of iterations at the start (``decreasing'' heuristic) or the end (``increasing'' heuristic) of the diffusion process (see \cref{fig:table1}). 
Additionally, \shortname{} supports adaptive allocations of forward passes; further discussions about adaptive algorithms can be found in the supplementary material.

\paragraph*{Reusing Solutions.}\label{sec:methods_reuse}

The convergence speed of fixed point solving meaningfully depends on the initial solution provided as input. A poor guess of the initial solution would make the convergence slower. Considering each timestep independently, the usual approach would be to initialize the fixed-point iteration of each timestep using the output of the (explicit) layers. However, considering sequential fixed point problems in the diffusion process, we can reuse the solution from the fixed point layer at the previous timestep as the initial solution for the next timestep. Formally, we can initialize the iteration in \cref{network} with $x^{* (t-1)}$ rather than with $x_{pre}^{(t)}$. 

The intuition for this idea is that adjacent timesteps of the diffusion process only differ by a small amount of noise, so their fixed point problems should be similar. Hence, the solutions of these problems should be close, and the solution of the previous timestep would be a good initialization for the current timestep. A similar idea was explored in \cite{deqflow}, which used fixed point networks for optical flow estimation.

\subsection{Stochastic Jacobian-Free Backpropagation} \label{ssec:backprop}

The final unresolved aspect of our method is how to train the network, i.e. how to effectively backpropagate through the fixed point solving layer. While early work on DEQs used expensive (Jacobian-based) implicit differentiation techniques~\cite{bai19deep}, recent work has found more success using approximate and inexpensive (Jacobian-free) gradients~\cite{fung22jacobian}. 

Concretely, this consists of first computing an approximate fixed point (either via iteration or Broyden's method) \textit{without storing any intermediate results for backpropagation}, and then taking a single additional fixed point step while storing intermediate results for backpropagation in the standard way.
During the backward pass, the gradient is only computed for the final step, and so it is referred to as a ``1-step gradient'' or Jacobian-Free Backpropagation (JFB). 
\footnote{We note that this process has been referred to in the literature by many names, including the 1-step gradient, phantom gradient, inexact gradient, and Jacobian-Free backpropagation.} 

Formally, this approximates the gradient of the loss $\fL$ with respect to the parameters $\theta$ of the implicit layer $f_{\text{fp}}$ with fixed point $x^{* (t)}$ by
\begin{align*}
  \ppx{\fL}{\theta} 
  &= \ppx{\fL}{x^{* (t)}} \left( I - \ppx{f_{\text{fp}}(x^{* (t)}, \tilde{x}^{(t)}, t) }{x^{* (t)}} \right) \ppx{f_{\text{fp}} (x^{* (t)}, \tilde{x}^{(t)}, t) }{\theta} \\
  &\approx \ppx{\fL}{x^{* (t)}} \ppx{f_{\text{fp}} (x^{* (t)}, \tilde{x}^{(t)}, t) }{\theta}
\end{align*}
The equality above is a consequence of the Implicit Function Theorem~\cite{babyrudin} and the approximation simply drops the inverse Jacobian term. This simplification is rigorously justified by Theorem 3.1 in~\cite{fung22jacobian} under appropriate assumptions.

Despite the simplicity of the 1-step gradient, we found that it performed poorly in large-scale experiments. To improve performance without sacrificing efficiency, we use a new stochastic approach to compute approximate gradients. 

Our approach (\cref{algo:FPDM-training-method}) samples random variables $n\sim U[0, N]$ and $m\sim U[1, M]$ at each training step. During the forward pass, we perform $n$ fixed point iterations without gradient to obtain an approximate solution,\footnote{By ``without gradient'', we mean that these iterations do not store any intermediate results for backpropagation, and they are not included in gradient computation during the backward pass.}, and then we perform $m$ additional iterations with gradient. 
During the backward pass, we backpropagate by unrolling only through the last $m$ iterations. 
The constants $N$ and $M$ are hyperparameters that refer to the maximum number of training iterations without and with gradient, respectively.
When sampling, the number of iterations used at each timestep is flexible and can be chosen independently of $N$ or $M$.
\begin{algorithm}
\caption{Stochastic Jacobian-Free Backpropagation}
\hspace*{\algorithmicindent} \textbf{Input} hidden states $x$, timestep $t$
\begin{algorithmic}[1]
  \Function{forward}{$x$} 
    \State $\tilde{x} \gets$ \Call{Proj}{$x$}  $\quad$\# input injection
    \For{$n$ iterations drawn uniformly from $0$ to $N$}
        \State $x \gets$ \Call{ForwardPassWithoutGrad}{$x, \tilde{x}, t$} 
    \EndFor
    \For{$m$ iterations drawn uniformly from $1$ to $M$}
        \State $x \gets$ \Call{ForwardPassWithGrad}{$x, \tilde{x}, t$} 
    \EndFor
    \State \Call{Backprop}{$loss, x$}
    \State \Return $x$ 
\EndFunction
\end{algorithmic}
\label{algo:FPDM-training-method}
\end{algorithm}
\vspace{-2mm}
Compared to the 1-step gradient, our method consumes more memory and compute because it backpropagates through multiple unrolled iterations rather than a single iteration. However, it is still drastically more efficient than either implicit differentiation or using traditional explicit networks, and it significantly outperforms the 1-step gradient in our experiments (\cref{tab:phantom_grad_08}).

\section{Experiments} \label{sec:experiments}

\subsection{Experimental Setup}

\paragraph*{Model}

The architecture of \shortname{} is based on the current state-of-the-art in generative image modeling, DiT-XL/2~\cite{william22scalable}, which serves as a strong baseline in our experiments. 
Adhering to the architecture in \cite{william22scalable}, we operate in latent space using the Variational Autoencoder from \cite{autoencoder,rombach2022high}. 
In addition, we have equipped both the baseline DiT and our \shortname{} with two advances from the recent diffusion literature: (1) training to predict velocity rather than noise~\cite{salimans2021progressive}, and (2) modifying the denoising schedule to have zero terminal signal-to-noise ratio~\cite{zerosnr}. 
We include these changes to show that our improvements are orthogonal to other improvements made in the diffusion literature. 

Our network consists of three sets of layers: pre-layers, an implicit fixed point layer, and post-layers. 
All layers have the same structure and 24M parameters, except the implicit layer has an additional projection for input injection. 
Through empirical analysis, we find that a single pre/post layer can achieve strong results (see \cref{sec:ablation_pre_post}). 
Consequently, the number of parameters in our full network is only 86M, markedly lower than 674M parameters in the standard DiT XL/2 model, which has 28 explicit layers. 
\paragraph*{Training}
We perform experiments on four diverse and popular datasets: Imagenet, CelebA-HQ, LSUN Church, and FFHQ. 
All experiments are performed at resolution 256. The ImageNet experiments are class-conditional, whereas those on other datasets are unconditional. 
For a fair comparison, we train our models and baseline DiT models for the same amount of time using the same computational resources. 
All models are trained on 8 NVIDIA V100 GPUs; the models for the primary experiments on ImageNet are trained for four days (equivalent to 400,000 DiT training steps), while those for the other datasets and for the ablation experiments are trained for one day (equivalent to 100,000 DiT steps). 
We train using Stochastic JFB with $M = N = 12$ and provide an analysis of this setup in \cref{sec:ablation_training}. 

The ImageNet experiments are class-conditional, whereas those on other datasets are unconditional. 
For ImageNet, following DiT, we train using class dropout of $0.1$, but we compute quantitative results without classifier-free guidance. We train with a total batch size of $512$ and learning rate $1$e$-4$. We use a linear diffusion noise schedule with $\beta_{\text{start}} = 0.0001$ and $\beta_{\text{end}} = 0.02$, modified to have zero terminal SNR~\cite{zerosnr}. We use $v$-prediction as also recommended by \cite{zerosnr}. Following DiT~\cite{william22scalable}, we learn the variance $\sigma$ along with the velocity $v$. 

Finally, with regard to the evaluations, all evaluations were performed using $50000$ images (FID-$50$K) except those in \cref{tab:adaptive} and \cref{fig:table5}, which were computed using $1000$ images due to computational constraints.

\subsection{Sampling Quality and Cost Evaluation}

To measure image quality, we employ the widely-used Frechet Inception Distance (FID) 50K metric~\cite{fidscore}. 
To measure the computational cost of sampling, previous studies on diffusion model sampling have counted the number of function evaluations (NFE)~\cite{lu2022dpm,karras2022elucidating,zhang2022fast}.
However, given the implicit nature of our model, a more granular approach is required. 
In our experiments, both implicit and explicit networks consist of transformer blocks with identical size and structure, so the computational cost of each sampling step is directly proportional to the number of transformer block forward passes executed; the total cost of sampling is the product of this amount and the total number of timesteps.\footnote{To be predise the implicit layer includes an extra projection for input injection, but this difference is negligible.} 
As a result, we quantify the sampling cost in terms of \textit{total transformer block forward passes}.
\footnote{
For example, sampling from a \shortname{} with one pre/post-layer and 26 fixed point iterations across $S$ timesteps requires the same amount of compute/time as a \shortname{} with two pre/post layers and $10$ iterations using $2S$ timesteps; this computation cost is also the same as that of a traditional DiT with 28 layers across $S$ timesteps. Formally, the sampling cost of \shortname{} is calculated by $(n_{\text{pre}} + n_{\text{iters}} + n_{\text{post}}) \cdot S$ where $n_{\text{pre}}$ and $n_{\text{post}}$ are the number of pre- and post-layers, $n_{\text{iters}}$ the number of fixed point iterations, and $S$ the number of sampling steps.
}

\begin{table}[t]
\small
\centering

\begin{tabular}{llcccc}
\toprule
\textit{\footnotesize Blocks} & \textit{\footnotesize Model} & \textit{\footnotesize FID} & \textit{\footnotesize FID} & \textit{\footnotesize Params.} & \textit{\footnotesize Training} \\
&  & \textit{\footnotesize(DDPM)} & \textit{\footnotesize(DDIM)} & \textit{\footnotesize } & \textit{\footnotesize Memory} \\
\midrule
\multirow{2}{*}{140} &  DiT        &  148.0         &  110.0         &  674M            &  25.2 GB          \\
                     &  \shortname &  \textbf{85.8} &  \textbf{33.9} &  \textbf{85M}    &  \textbf{10.2 GB} \\ \midrule 
\multirow{2}{*}{280} &  DiT        &  80.9          &  35.2          &  674M            &  25.2 GB          \\
                     &  \shortname &  \textbf{43.3} &  \textbf{22.4} &  \textbf{85M}    &  \textbf{10.2 GB} \\ \midrule
\multirow{2}{*}{560} &  DiT        &  37.9          &  \textbf{16.5} &  674M            &  25.2 GB          \\
                     &  \shortname &  \textbf{26.1} &  19.6          &  \textbf{85M}    &  \textbf{10.2 GB} \\ \bottomrule
\end{tabular}

\caption{\textbf{Quantitative Results on ImageNet.} Despite having 87$\%$ fewer parameters and using 60\% less memory during training, \shortname{} outperforms DiT~\cite{william22scalable} at 140 and 280 transformer block forward passes and achieves comparable performance at 560 passes. 
}%
\vspace{-3mm}
\label{tab:quantitative_01}
\end{table}

\begin{table}[th!]
\small
\centering

\begin{tabular}{llr|llr}
\toprule
\textit{\small Dataset} & \textit{\small Model} & \textit{\small FID} & \textit{\small Dataset} & \textit{\small Model} & \textit{\small FID} \\
\midrule
{\footnotesize CelebA-HQ}  &  DiT        &  65.2          &  {\footnotesize FFHQ}     &  DiT                   &  58.1           \\
                           &  \shortname &  \textbf{11.1} &                           &  \shortname            &  \textbf{18.2}  \\ \midrule
{\footnotesize LSUN-Church} &  DiT        &  65.6          &  {\footnotesize ImageNet} &  DiT                   &  80.9           \\
                           &  \shortname &  \textbf{22.7} &                           &  \shortname            &  \textbf{43.3}  \\ \bottomrule
\end{tabular}

\caption{\textbf{Quantitative Results Across Four Datasets.} \shortname{} consistently outperforms DiT~\cite{william22scalable} on CelebA-HQ, FFHQ, LSUN-Church, and Imagenet with 280 transformer block forward passes. 
All models are trained and evaluated at resolution 256px using the same amount of compute and identical hyperparameters.
}%
\vspace{-3mm}
\label{tab:datasets_02}
\end{table}
\begin{figure}
\centering
\hspace{-5mm}\includegraphics[width=0.48\textwidth]{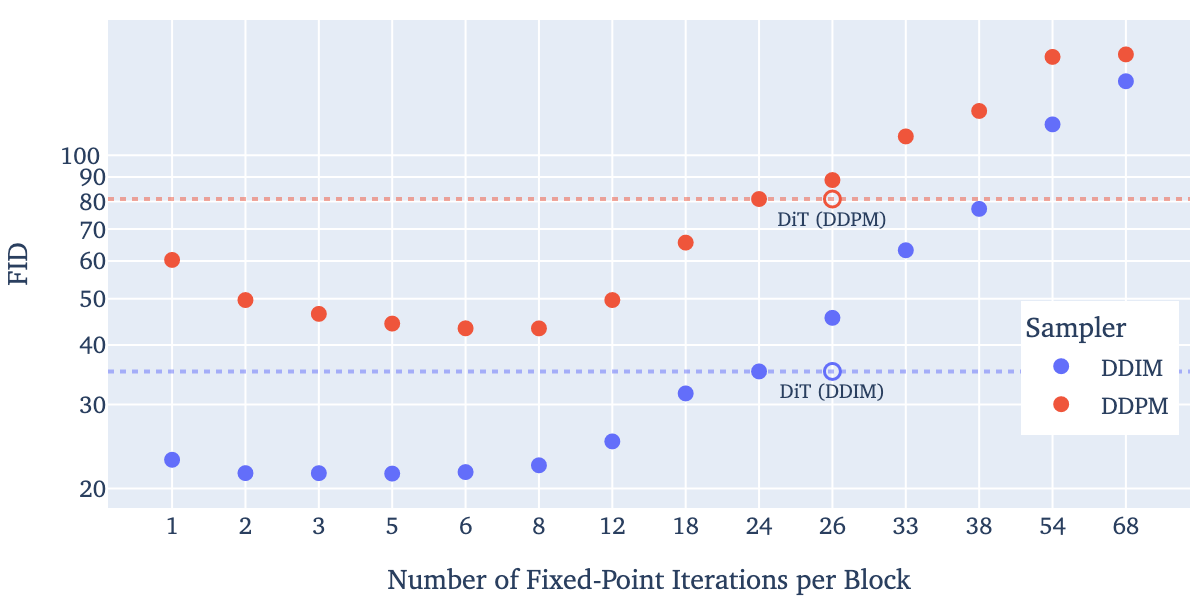}
\caption{\textbf{Timestep smoothing significantly improves performance.} Given the same amount of sampling compute (280 transformer blocks), \shortname{} enables us to allocate computation among more or fewer diffusion timesteps, creating a tradeoff between the number of fixed-point solving iterations per timestep and the number of timesteps in the diffusion process (See \cref{sec:methods_smoothing}). 
Here we explore the performance of our model on ImageNet with fixed point iterations ranging from 1 iteration (across 93 timesteps) to 68 iterations (across 4 timesteps). Each timestep also has 1 pre- and post-layer, so sampling with $k$ iterations utilizes $k+2$ blocks of compute per timestep.
The circle and dashed lines show the performance of the baseline DiT-XL/2 model with 28 layers, which in our formulation corresponds to smoothing over 26 iterations. 
Although our model is slightly worse than DiT at 26 iterations, it significantly outperforms DiT when smoothed across more timesteps, demonstrating the effectiveness of timestep smoothing.
}
\label{fig:iterations_timesteps_03}
\end{figure}
\begin{figure*}[t]
\centering
\includegraphics[width=\textwidth]{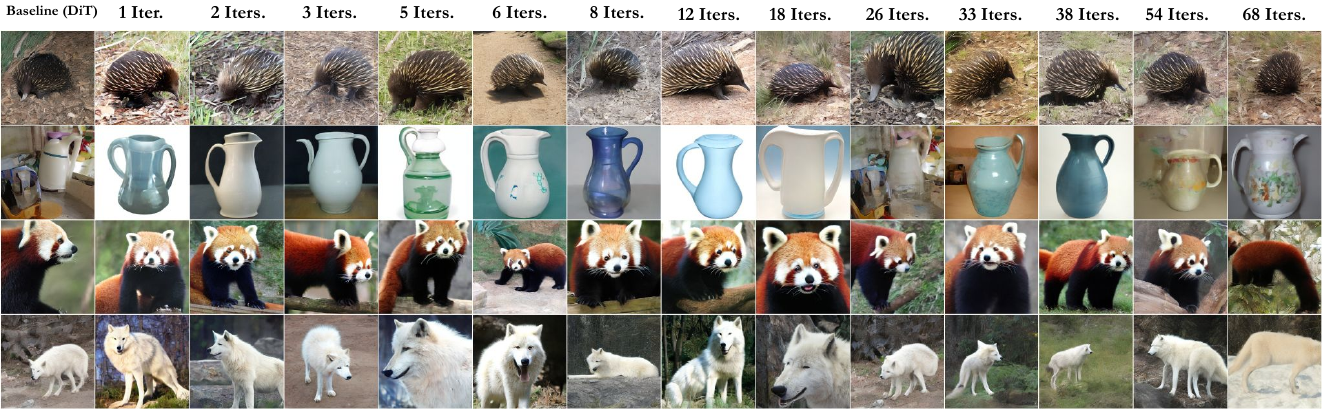}
\caption{\textbf{Qualitative Results for Smoothing Computation Across Timesteps}. We show visual results of \shortname{} using different numbers of fixed point solving iterations, while keeping the total amount of sampling compute fixed (560 transformer blocks). Our method demonstrates similar performance compared to the baseline with 20 to 30 iterations per timestep and superior generation quality with 4 to 8 iterations, as reflected quantitatively in \cref{fig:iterations_timesteps_03}.
}%
\label{fig:qualitative_iterations_timesteps}
\end{figure*}

\subsection{Results}

In \cref{tab:quantitative_01}, we first present a quantitative comparison of our proposed \shortname{} against the baseline DiT, under different amounts of sampling compute. 
Notably, given 140 and 280 transformer block forward passes, our best model significantly outperforms DiT, with the widest performance gap given the most limited compute. 
Our method's improvements are orthogonal to those gained from using better samplers; our model effectively lowers the FID score with both DDIM and DDPM.
At 560 forward passes, our method outperforms DiT with DDPM but not DDIM, and for more than 560 it is outperformed by DiT. 
Note that the number of parameters in \shortname{} is only 12.9\% of that in DiT, and it consumes 60\% less memory during training (reducing memory from $25.2$ GB to only $10.2$ GB at a batch size of $64$). 

\cref{tab:datasets_02} extends the comparison between \shortname{} and DiT to three additional image datasets: FFHQ, CelebA-HQ, and LSUN-Church. Our findings are consistent across these datasets, with \shortname{} markedly improving the FID score despite being nearly one-tenth the size of DiT.

\cref{fig:splash} shows qualitative results of our model compared to DiT. 
All images are computed using the same random seeds and classes using a classifier-free guidance scale 4.0 and $560$ transformer block forward passes (20 timesteps for DiT). 
\shortname{} uses $8$ fixed point iterations per block with timestep smoothing.
Our model produces sharper images with more detail, likely due to its ability to spread the computation among timesteps, as discussed in \cref{sec:discussion_smoothing}. 

\begin{figure}[ht]
\centering
\hspace{-5mm}\begin{subfigure}{0.48\textwidth}
\centering
\includegraphics[width=\linewidth]{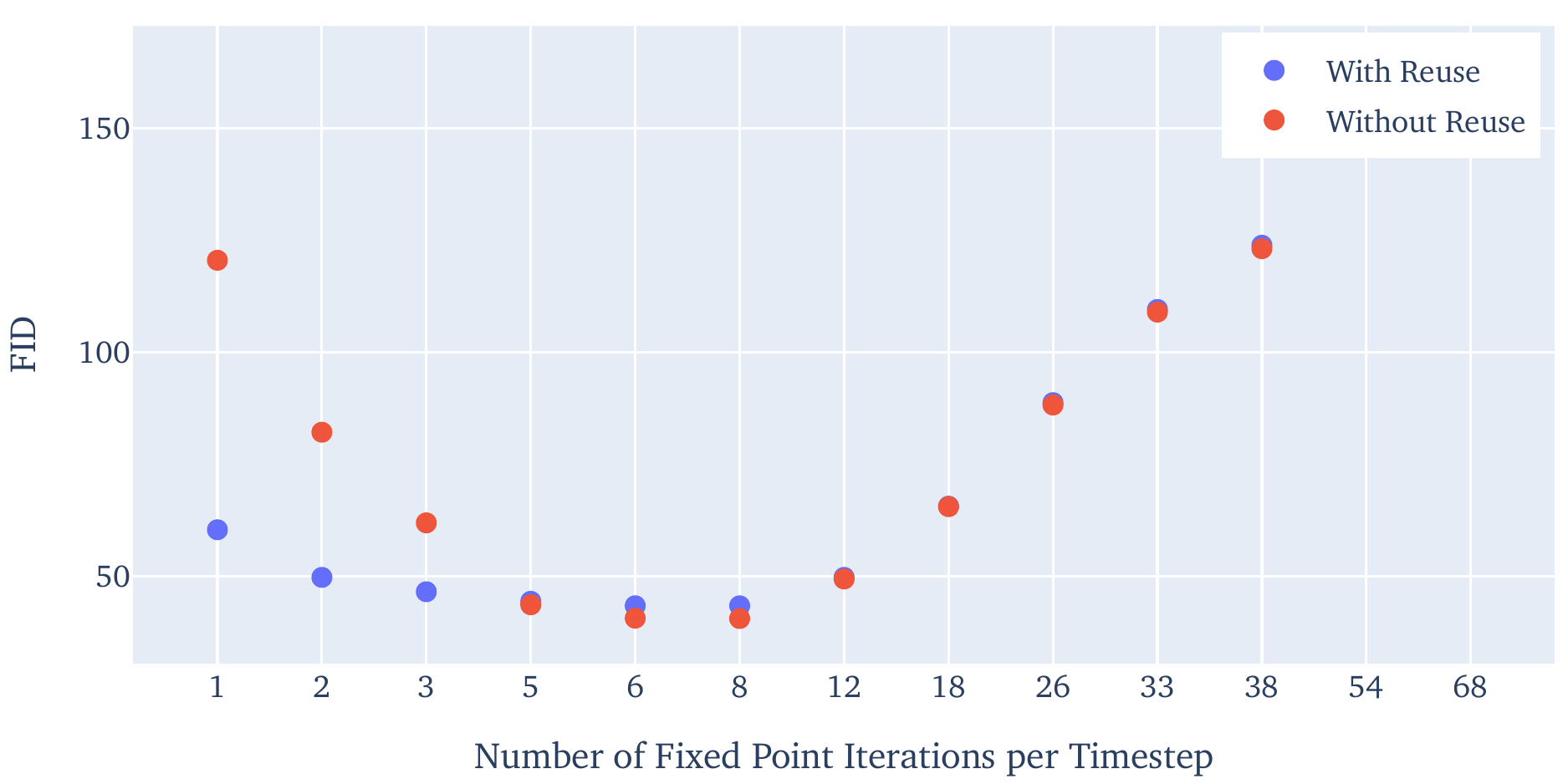}
\caption{
Performance improvement from reusing solutions across timesteps.
}%
\label{fig:reuse_comparison}
\end{subfigure}
\vspace{1em} %
\hspace{-5mm}\begin{subfigure}{0.48\textwidth}
    \centering
    \includegraphics[width=\linewidth]{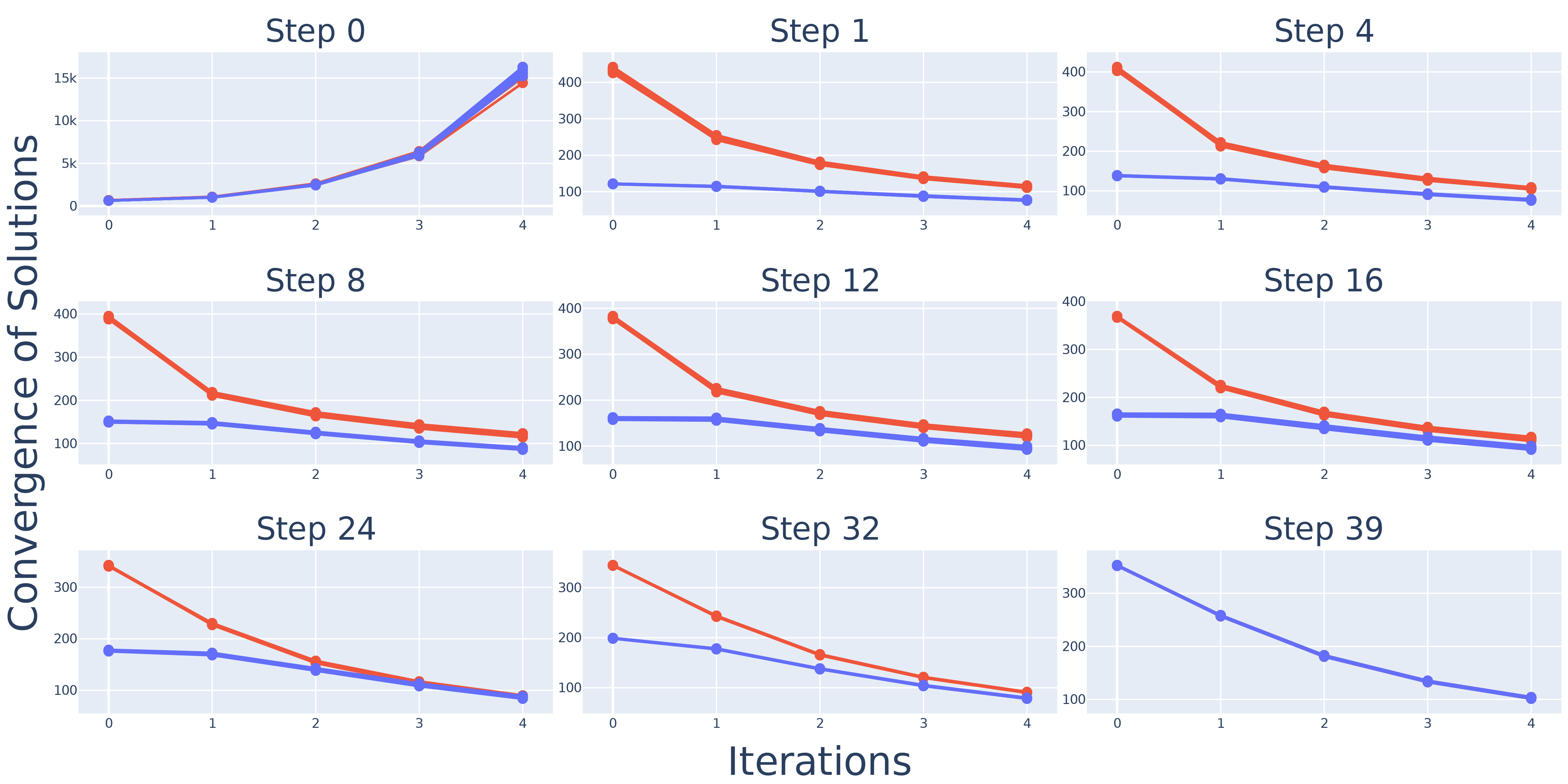}
    \caption{\footnotesize Convergence of f.p. iteration for nine timesteps, with and without reuse.}
    \label{fig:table5a}
\end{subfigure}
\vspace{1em} %
\hspace{-5mm}\begin{subfigure}{0.48\textwidth}
    \centering
    \includegraphics[width=\linewidth]{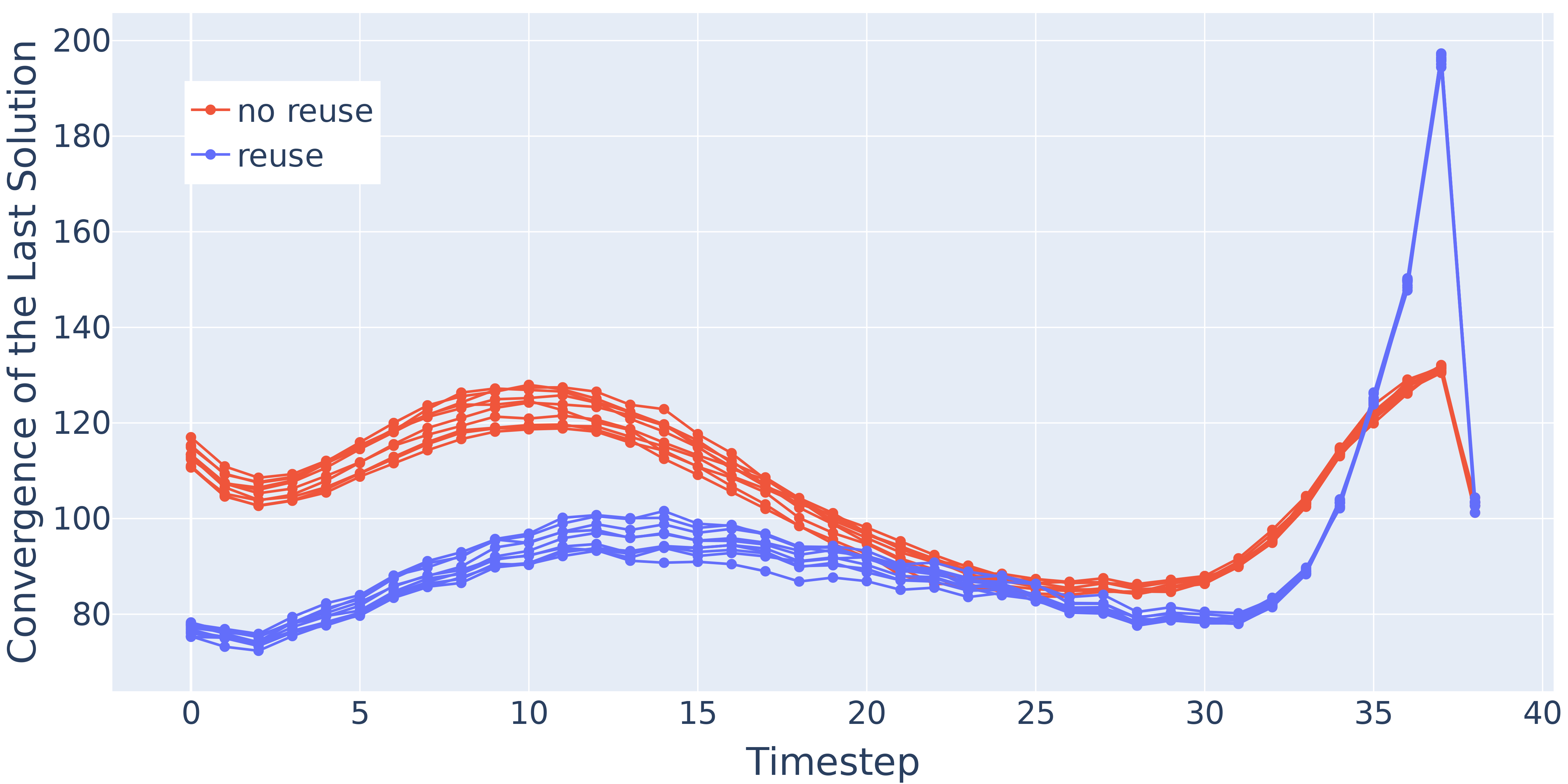}
    \caption{\footnotesize Convergence at iteration across timesteps, with and without reuse.}
    \label{fig:table5b}
\end{subfigure}
\vspace{-5mm}
\caption{\textbf{The Impact of Reuse on Fixed Point Accuracy.} In (a), we examine the performance of sampling with and without reusing solutions for different numbers of iterations per timestep; it considerably helps when using a small number of iterations per timestep. 
In (b) and (c), we examine the convergence of individual timesteps. We see that reuse delivers particularly large benefits for smaller (less-noisy) timesteps. Note that these plots contain 10 lines as they are plotted for 10 random batches of 32 images from ImageNet.}
\vspace{-5mm}
\label{fig:table5}
\end{figure}

\begin{table}[t]
\small
\centering

\begin{tabular}{lrrrr}
\toprule
\textit{\small Train Iters.} ($M$, $N$) & \textit{3} & \textit{6} & \textit{12} & \textit{24} \\
\midrule
\textit{\small FID} & \textbf{43.0} & 43.2 & 61.5 & 567.6 \\
\bottomrule
\end{tabular}   

\caption{\textbf{Performance For Varying Numbers of Fixed Point Iterations in Training.} This table compares various choices of $M$ and $N$ values in \cref{algo:FPDM-training-method}. They represent a tradeoff between training speed and fixed point convergence accuracy. Results indicate that the optimal values for $M$ and $N$ range from 3 to 6.%
}
\vspace{-4mm}
\label{tab:training_iters_07}
\end{table}
\begin{table}[t]
\small
\centering

\begin{tabular}{lrr}
\toprule
\textit{\footnotesize Train iters without grad ($N$)} & \textit{\small Method}         &   \textit{\small FID} \\ \midrule
             & JFB (1-Step Grad)   &   567.6           \\
\textit{6}   & Multi-Step JFB        &   48.2            \\ 
             & Stochastic JFB      &   \textbf{43.2}   \\ \midrule
             & JFB (1-Step Grad)   &   567.6           \\
\textit{12}  & Multi-Step JFB        &   79.9            \\ 
             & Stochastic JFB      &   \textbf{61.5}   \\ \midrule
\end{tabular}

\caption{\textbf{Performance of Stochastic Jacobian-Free Backpropagation (S-JFB) compared to JFB (1-step gradient).} We find that 1-step gradient, the most common method for training DEQ~\cite{bai19deep} models, struggles to optimize models on the large-scale ImageNet dataset, whereas a multi-step version of it performs well and our stochastic multi-step version performs (S-JFB) even better. The 1-step gradient always unrolls with gradient through a single iteration ($M=1$) of fixed point solving, whereas the stochastic version unrolls though $m \sim U(1, M)$ iterations for $M=12$.} 
\vspace{-1mm}
\label{tab:phantom_grad_08}
\end{table}

\begin{figure}
\centering
\includegraphics[width=0.480\textwidth]{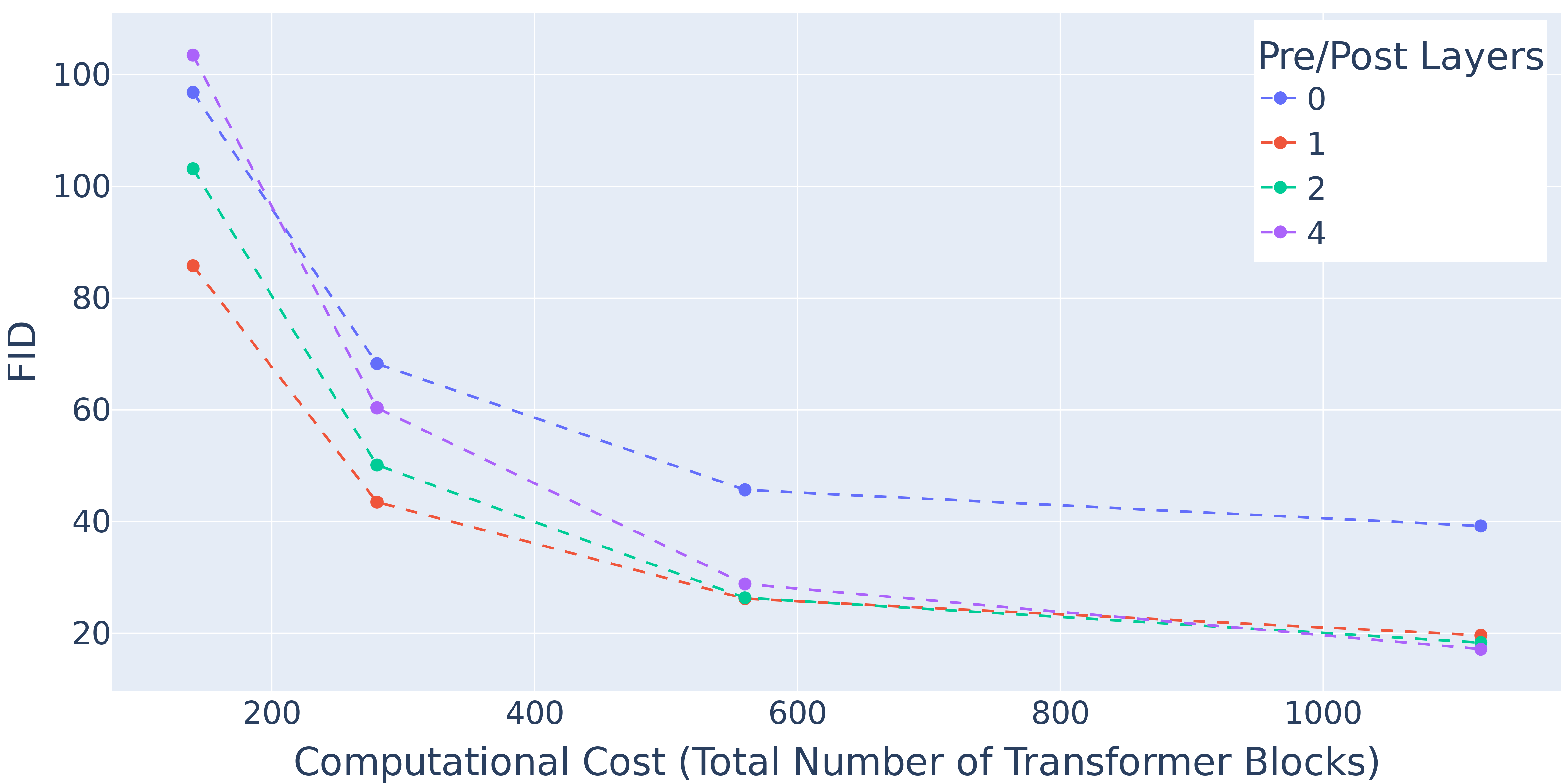}
\caption{\textbf{Performance of Different Number of Pre/Post Layers.} We find that using at least one pre/post layer is always better than none; fewer explicit layers perform better on small compute budgets, whereas more explicit layers can better leverage large budgets.}%
\label{fig:pre_post_09}
\vspace{-4mm}%
\end{figure}
\begin{table}[t]
\small
\centering

\begin{tabular}{lllllll}
\toprule
{\footnotesize Iters. per Step} & 3 & 5 & 6 & 8 & 12 & 26 \\
\midrule

\textit{Constant} & 48.0 & 45.8 & 46.6 & 47.3 & 48.5 & 62.5 \\
\textit{Decreasing} & 48.0 & 46.3 & 47.3 & 48.3 & 49.1 & 63.2 \\
\textit{Increasing} & $\textbf{46.7}$ & $\textbf{44.8}$ & $\textbf{45.9}$ & $\textbf{45.6}$ & $\textbf{48.0}$ & $\textbf{61.7}$ \\
\bottomrule
\end{tabular}
\caption{
\textbf{Performance of Iteration Allocation Heuristics.} \textit{Constant} uses a fixed iteration count per diffusion timestep, while \textit{Increasing} and \textit{Decreasing} vary their iteration counts linearly with respect to the timestep.
}
\vspace{-2mm}
\label{tab:adaptive}
\end{table}

\section{Analysis and Ablation Studies} \label{sec:ablations}

\subsection{Smoothing Computation Across Timesteps}\label{sec:discussion_smoothing}
In \cref{fig:iterations_timesteps_03} and \ref{fig:qualitative_iterations_timesteps}, we examine the effect of smoothing timesteps as described in \cref{sec:methods_smoothing}. 
We sample across a range of fixed point iterations and timesteps, keeping the total sampling cost (i.e. the total number of transformer block forward passes) constant. 
This explores the trade-off between the convergence of the fixed point iteration at each timestep and the discretization error of the larger diffusion process. 

Balancing the number of iterations and timesteps is key to obtaining optimal performance. 
Intuitively, when using very few iterations per timestep, the process fails to converge adequately at each step, and the resulting error compounds. 
Conversely, allocating too many iterations to too few timesteps results in unnecessary computation on already converged solving iterations, resulting in discretization errors arising from larger gaps between timesteps.
An ideal strategy involves using just enough fixed-point iterations to achieve a satisfactory solution, thereby maximizing the number of possible timesteps. 
For instance, with 280 transformer block forward passes, we see that the optimal range lies between 4 and 8 iterations per timestep. 

\subsection{Reallocating Computation Across Timesteps}

In \cref{tab:adaptive}, the increasing heuristic outperforms the contant and decreasing ones; allocating resources more toward the later stages of the denoising process improves generation quality and detail. 
Note that such flexibility in resource allocation is a novel feature of \shortname, not possible in previous explicit diffusion models.

\subsection{Reusing Solutions}

As described in \cref{sec:methods_reuse}, we explore reusing the fixed point solution from each timestep to initialize the subsequent step. 
In \cref{fig:reuse_comparison}, we see that reusing makes a big difference when 
performing a small number of iterations per timestep
and a negligible difference when 
performing many iterations per timestep. 
Intuitively, reusing solutions reduces the number of iterations needed at each timestep, so it improves the performance when the number of iterations is severely limited. 
\cref{fig:table5a} and \cref{fig:table5b} illustrate the functionality of reusing by examining at the individual timestep level. For each timestep $t$, we use the difference between the last two fixed point iterations, $\delta_t$, as an indicator for convergence.
Reusing decreases $\delta_t$ for all timesteps except a few noisiest steps, and reusing is most effective at less noisy timesteps. 
This observation aligns with our intuition: Adjacent timesteps with less noise tend to have highly similar corresponding fixed point systems, where reusing is more effective.

\subsection{Pre/Post Layers}
\label{sec:ablation_pre_post}

One of the many ways \shortname{} differs from prior work on DEQs is that we include explicit layers before and after the implicit fixed point layer. In \cref{fig:pre_post_09}, we conduct an ablation analysis, training networks with $0$, $1$, $2$, and $4$ pre/post layers. 
We see that using at least $1$ explicit pre/post layer is always better than $0$.
For small compute budgets it is optimal to use $1$ pre/post layer, and for larger budgets it is optimal to use $2$ or $4$. 
Broadly, we observe that using more explicit layers limits flexibility thereby reducing performance at lower compute budgets, but improves performance at higher budgets.

\subsection{Training Method}
\label{sec:ablation_training}
In \cref{tab:training_iters_07}, we compare versions of Stochastic Jacobian-Free Backpropagation with different values of $M$ and $N$, the upper bounds on the number of training iterations with and without gradient. $M$ and $N$ reflect a training-time tradeoff between speed and fixed point convergence. 
As $M$ and $N$ increase, each training step contains more transformer block forward and backward passes on average; the fixed point approximations become more accurate, but each step consumes more time and memory. 
We find the optimal values of $M$ and $N$ are quite low: $3$ or $6$. 
Using too many iterations (e.g. $24$) is detrimental as it slow down training. 

In \cref{tab:phantom_grad_08}, we compare to JFB (also called $1$-step gradient), which has been used in prior work on DEQs~\cite{fung22jacobian}, and a multi-step variant of it. We find that training with multiple steps is essential to obtaining good results, and that using a stochastic number of steps delivers further improvements.

\subsection{Limitations}
The primary limitation of our model is that it performs worse than the fully-explicit DiT model when sampling computation and time are not constrained. 
The performance gains from our model in resource-constrained settings stem largely from smoothing and reusing, but in scenarios with saturated timesteps and iterations, the efficacy of these techniques is reduced. 
In such cases, our network resembles a transformer with weight sharing~\cite{reid2021subformer,lan2019albert}, which underperform vanilla transformers. 
Hence, we do not expect to match the performance of DiT, which has 8$\times$ more parameters, when sampling with an unlimited amount of resources. 

\section{Conclusions}
We introduce \shortname{}, a pioneering diffusion model characterized by fixed point implicit layers. 
Compared to traditional Diffusion Transformers (DiT), \shortname{} significantly reduces model size and memory usage. 
In the context of diffusion sampling, \shortname{} enables us to develop new techniques such as solution reusing and timestep smoothing, which give \shortname{} enhanced flexibility in computational allocation during inference. 
This flexibility makes it particularly effective in scenarios where computational resources are constrained. 
Future work could explore new ways of leveraging this flexibility as well as scaling to larger datasets such as LAION-5B~\cite{laion5b}. 

{
    \small
    \bibliographystyle{ieeenat_fullname}
    \bibliography{refs_diffusion,refs,refs_xingjian_added}
}

\clearpage
\setcounter{page}{1}
\maketitlesupplementary

\appendix

\section{Diffusion Models}

As stated in \cref{sec:methods_fpdm}, we provide an overview of diffusion models here in case any readers are not familiar with diffusion models. 

Diffusion denoising probabilistic models add noise to a data sample \(X_0\) drawn from a target data distribution \(q(X_0)\). This noising process is executed in a series of steps, where each step adds a specific quantity of noise controlled by a variance schedule \(\{\beta_t\}_{t=0}^{T}\).
At each step, the new data sample \(X_t\) is generated from the previous one \(X_{t-1}\) according to the distribution \(q(X_t|X_{t-1}) = \mathcal{N}(X_t; \sqrt{1-\beta_t}X_{t-1}, \beta_t \mathbf{I})\). 
The reverse diffusion process, or generative process, starts with a noisy sample from \(q(X_T)\) $\sim \mathcal{N}(0, 1)$ and aims to iteratively remove the noise to recover a sample from the original distribution \(q(X_0)\). 
This reverse process is learned by a neural network, approximating the distribution \(q(X_{t-1}|X_t)\) as \(s_{\theta}(X_{t-1}|X_t) \approx q(X_{t-1}|X_t)\).

\section{Additional Qualitative Examples}

\begin{figure*}
\centering
\includegraphics[width=\textwidth]{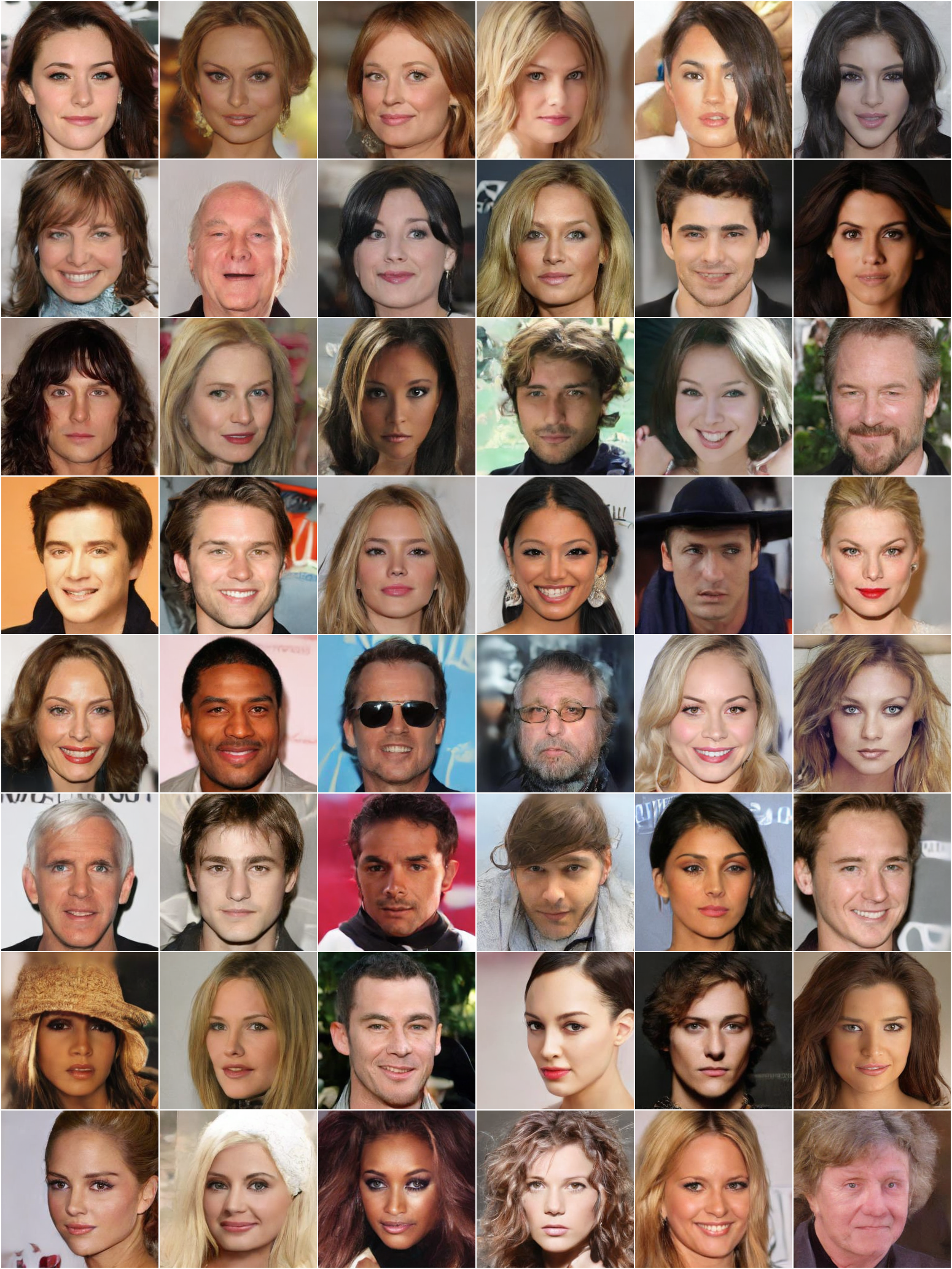}
\caption{\textbf{Additional qualitative examples on CelebA-HQ.} Examples are sampled using the DDPM sampler with 560 transformer block forward passes at resolution $256$px. These images are not cherry-picked. }%
\label{fig:additional_samples_celebahq}
\end{figure*}

\begin{figure*}
\centering
\includegraphics[width=\textwidth]{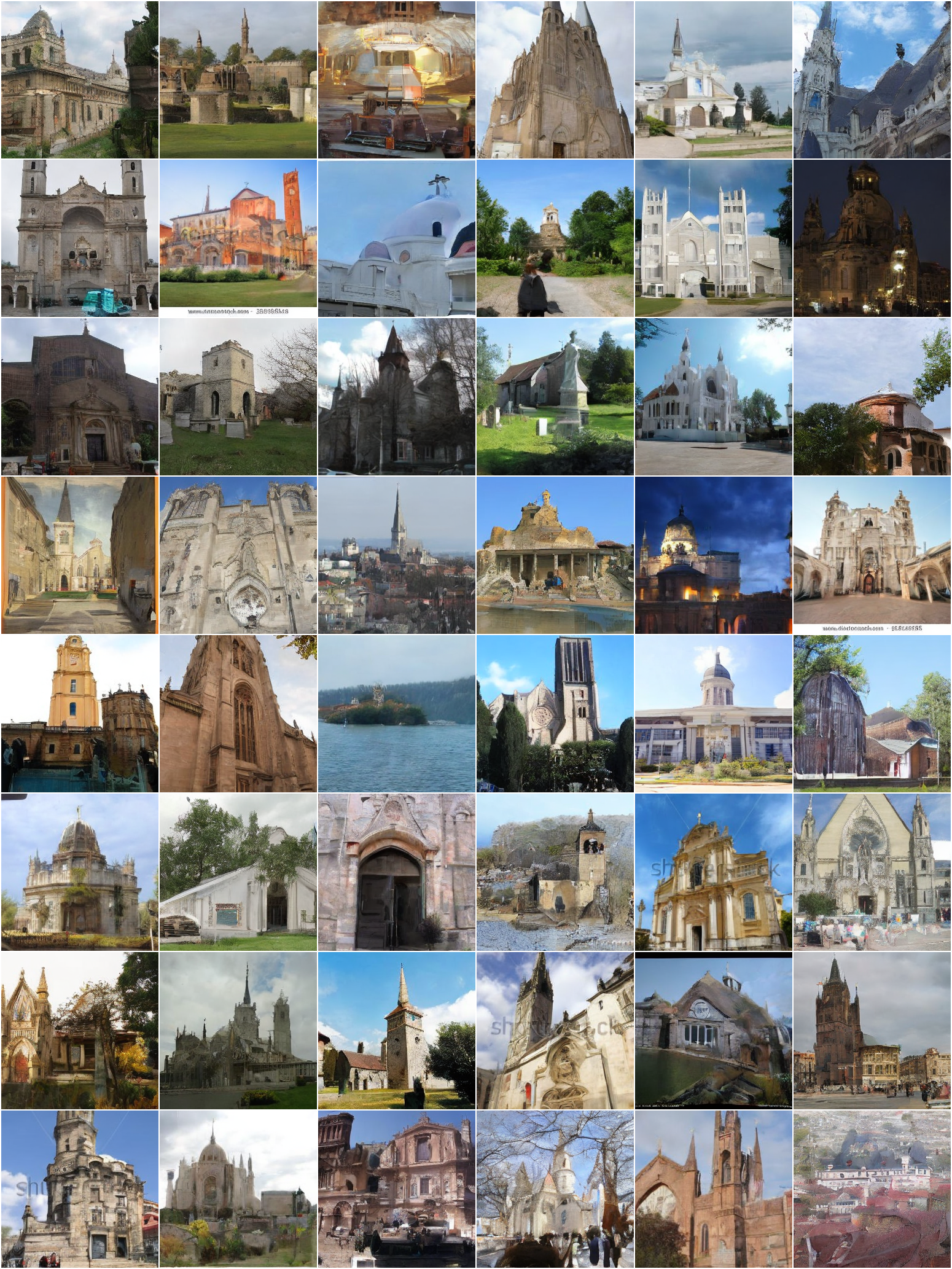}
\caption{\textbf{Additional qualitative examples on LSUN Church.} Examples are sampled using the DDPM sampler with 560 transformer block forward passes at resolution $256$px. These images are not cherry-picked. }%
\label{fig:additional_samples_lsun_church}
\end{figure*}

\begin{figure*}
\centering
\includegraphics[width=\textwidth]{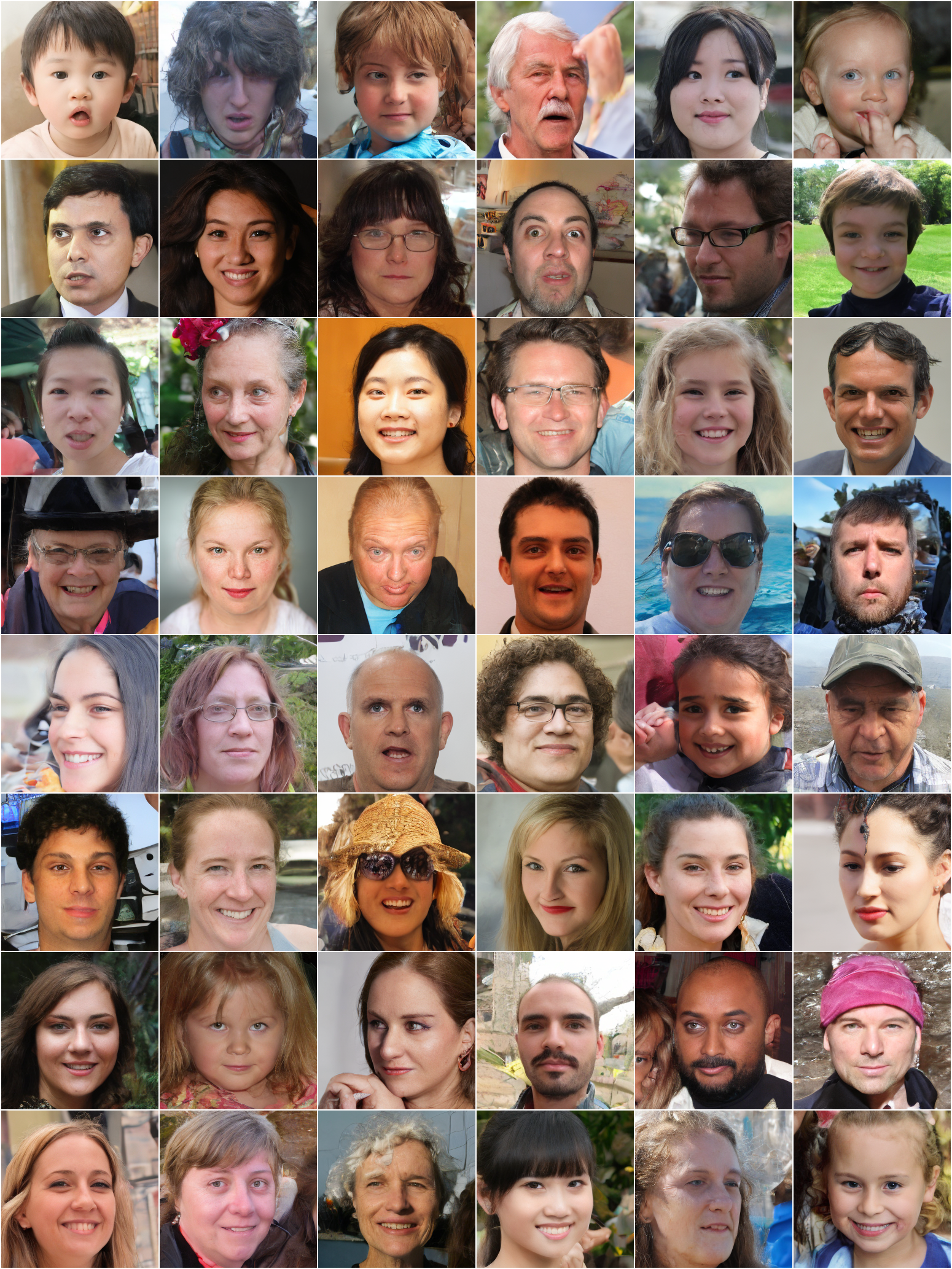}
\caption{\textbf{Additional qualitative examples on FFHQ.} Examples are sampled using the DDPM sampler with 560 transformer block forward passes at resolution $256$px. These images are not cherry-picked. }%
\label{fig:additional_samples_ffhq}
\end{figure*}

\begin{figure*}
\centering
\includegraphics[width=\textwidth]{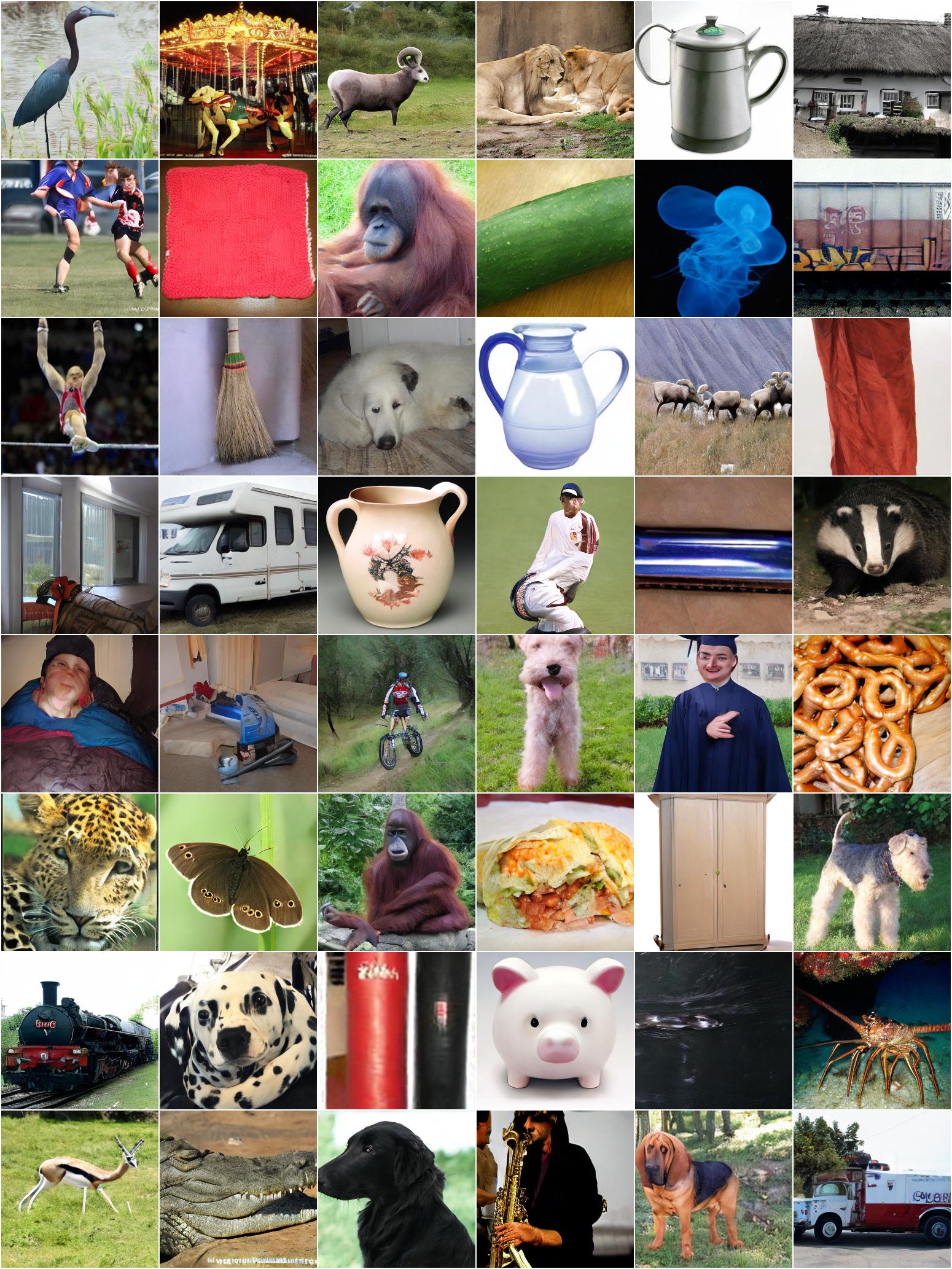}
\caption{\textbf{Additional qualitative examples on ImageNet.} Examples are sampled using the DDPM sampler with 560 transformer block forward passes at resolution $256$px. These images are not cherry-picked. }%
\label{fig:additional_samples_imagenet}
\end{figure*}

\begin{figure*}
\centering
\includegraphics[width=\textwidth]{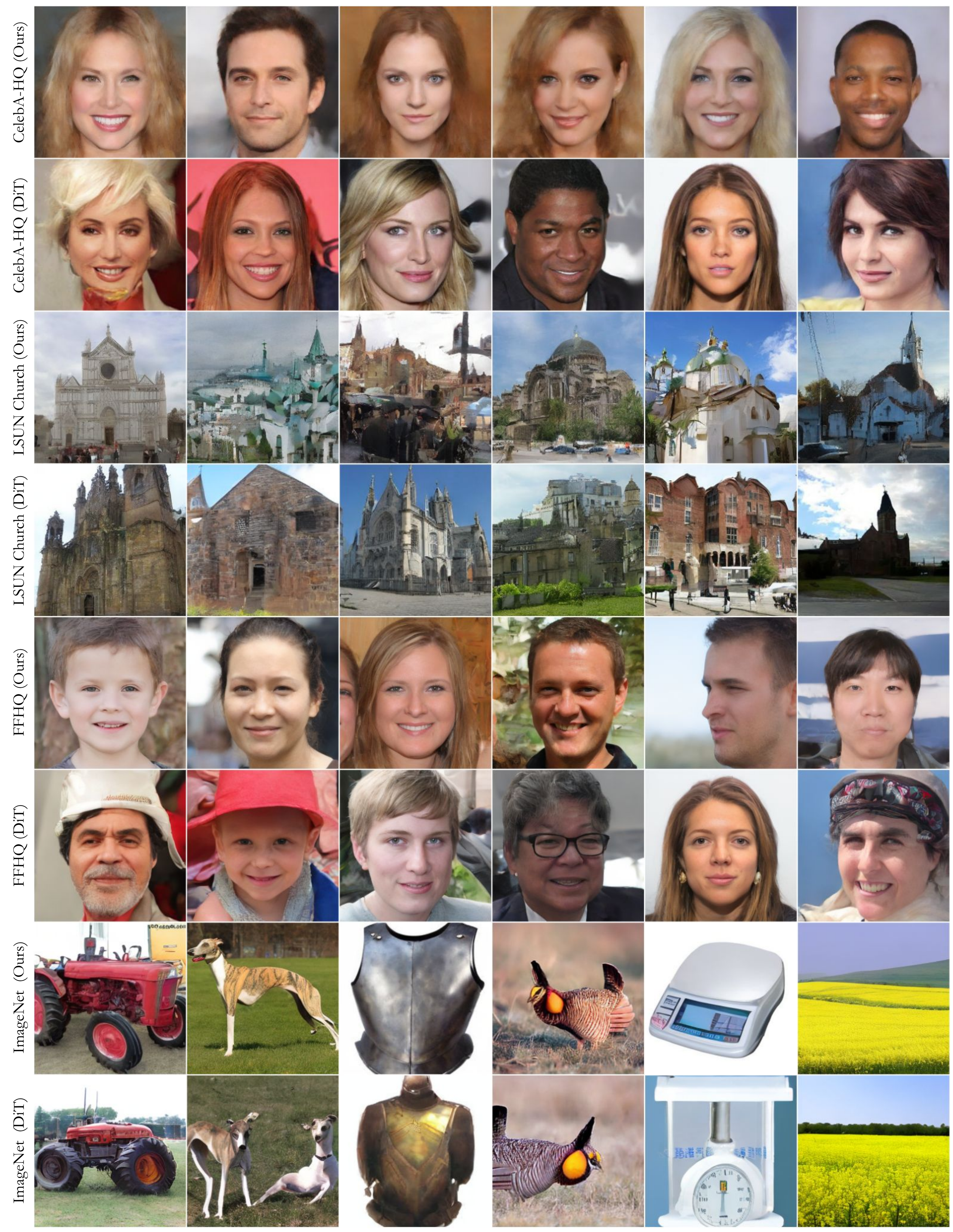}
\caption{\textbf{Additional qualitative comparison with DiT.} We show examples on CelebA-HQ, LSUN Church, FFHQ, and ImageNet. All images are sampled using the DDPM sampler with 560 transformer block forward passes at resolution $256$px. The images are not cherry-picked. }%
\label{fig:additional_samples_comparison}
\end{figure*}

We provide examples on CelebA-HQ, LSUN Church, FFHQ, and ImageNet in \cref{fig:additional_samples_celebahq,fig:additional_samples_lsun_church,fig:additional_samples_ffhq,fig:additional_samples_imagenet}. 
For each dataset, we sample $48$ images using DDPM with 560 transformer block forward passes at resolution $256$px. 
Note that the images are not cherry-picked. 
We also provide additional qualitative comparisons with DiT in \cref{fig:additional_samples_comparison}.

\section{Description of an Adaptive Allocation Algorithm}
\shortname{} allows for the adjustment of solution accuracy at different stages of the denoising process. 
As noted in \cref{sec:methods_fpdm}, in addition to implementing straightforward heuristics such as ``increasing'' and ``decreasing'', it supports using adaptive algorithms to allocate the forward passes across timesteps. 
We leave an in-depth investigation of adaptive algorithms to future work, but we give an example below to demonstrate how one such algorithm could work. 

We start by considering $\theta_t$, the difference between the last two solving solutions, as a metric of solution quality at each step. This aligns with our observation in \cref{fig:table5a}, where $\theta_t$ decreases as more fixed point iterations (i.e. forward passes) are applied. Then a simple adaptive algorithm could be to simply set an error threshold $\Theta$ and iterate the fixed point iteration process at each timestep $t$ continue until $\theta_t$ falls below $\Theta$. 
Then the global threshold $\Theta$ controls the number of forward passes. 

The only question left would be how to choose $\Theta$ to match a given computational budget (i.e. a number of forward passes). For this, an online binary probing scheme can be employed: use binary search to select a $\Theta'$, on which we perform inference for one batch of images. If the number of forward passes used to meet the $\Theta'$ threshold exceeds our budget, we increase $\Theta'$ in subsequent iterations; conversely, if the number is below our budget, we decrease $\Theta'$. Note that only constant time of probing is needed to find a sufficiently good threshold at the beginning of the inference. This computational cost would be negligible, especially when sampling many batches of images.

\end{document}